\documentclass[11pt,a4paper]{article}

\usepackage[utf8]{inputenc}
\usepackage[T1]{fontenc}
\usepackage{times}
\usepackage{microtype}

\usepackage[a4paper, margin=2.5cm, top=3cm, bottom=3cm]{geometry}

\usepackage{booktabs}
\usepackage{tabularx}
\usepackage{array}

\usepackage{amsmath}
\usepackage{amssymb}

\usepackage{cite}

\usepackage[hidelinks, breaklinks=true]{hyperref}
\usepackage{url}
\usepackage{fancyhdr}
\usepackage{titlesec}
\titleformat{\section}{\large\bfseries}{\thesection}{1em}{}
\titleformat{\subsection}{\normalsize\bfseries}{\thesubsection}{1em}{}
\titlespacing*{\section}{0pt}{12pt plus 2pt}{4pt}
\titlespacing*{\subsection}{0pt}{8pt plus 1pt}{2pt}

\title{\textbf{Per-Entity Bias Mapping for AI Visibility:}\\[4pt]
Why Brand Mentions Require Entity-Specific Calibration}

\author{Zolt\'an Varga\\[2pt]
  Neural Awareness\\
  Budapest, Hungary\\[2pt]
  \small\texttt{vargazoltan.ai}}

\date{v5 --- Ghost Cartography Edition (2026-05-27)\\[4pt]
  \small Zenodo preprint: \href{https://doi.org/10.5281/zenodo.20308957}{\texttt{10.5281/zenodo.20308957}}
  \quad $\cdot$ \quad
  v5 record: \href{https://doi.org/10.5281/zenodo.20419277}{\texttt{10.5281/zenodo.20419277}}}

\begin{document}

\maketitle
\thispagestyle{fancy}

\begin{abstract}
AI-mediated answer systems increasingly determine how brands and organizations are
represented to users. Existing approaches reduce visibility to mention rate or citation
frequency. This paper argues that aggregate metrics are insufficient because entities
exhibit systematically different AI visibility error profiles.

We identify three failure modes. First, underrepresented entities suffer invisibility
due to weak structured data and limited knowledge graph presence. Second, large familiar
entities suffer disproportionately from false attribution---the \emph{Brand Hallucination
Paradox}: model familiarity creates a stronger surface for plausible but incorrect
completions. Third, Central and Eastern European entities face a structural \emph{CEE
Entity Infrastructure Gap} across three layers: knowledge graph absence, NER deficits
for morphologically complex languages, and entity linking difficulty---producing AI
visibility deficits content optimization cannot resolve. A fourth dimension: the
\emph{Parametric-Retrieval Lag Asymmetry}, where retrieval-augmented systems update
within days while parametric memory updates only at retraining intervals of twelve to
twenty-four months, producing divergent representations across platforms.

We introduce Per-Entity Bias Mapping (PEBM): a ten-dimensional framework distinguishing
raw from verified mentions across retrieval inclusion, hallucination rate, citation
fidelity, source authority, and parametric-retrieval lag. A full-scale empirical study
($n{=}100$ Hungarian B2B entities, 1{,}400 probe runs, 2{,}062 sources, non-AI HTTP
verification) finds Tier~1 high-salience brands produce 52.69\%
fabricated citations (95\%~CI [49.76\%, 55.61\%]) versus 37.87\% for Tier~3
low-salience entities (95\%~CI [34.84\%, 41.00\%])
($+$14.82~pp; $\chi^2{=}45.326$, $p{=}1.67{\times}10^{-11}$, Cohen's $h{=}0.299$),
strongly supporting the Brand Hallucination Paradox. Regulatory-framed queries
elevate fabrication to 56.77\% (FDR $q{<}0.001$), versus 37.59\% for factual
queries---a $+19.2$~pp framing delta that constitutes a passive adversarial
attack surface requiring no model access. We further identify
\emph{rejection-induced confabulation escalation}: repeated non-acceptance of AI
responses in prompt chains systematically elevates fabrication via five converging
mechanisms, with the counterintuitive implication that agentic quality filters
function as hallucination accelerators in compliance query contexts. The framework
applies entity-level calibration from algorithmic fairness research to commercial AI
representation and provides open instruments for entity-level AI audit practice. We introduce
\emph{ghost cartography} as a unifying mechanism for entity-level AI misrepresentation: when
entities occupy sparse latent regions, models produce confident, evidence-like output interpolated
from neighboring dense regions. This yields a two-dimensional confabulation space in which the
Brand Hallucination Paradox captures fabricated presence (Type~3) and Parametric-Retrieval Lag
Asymmetry captures frozen representation (Type~4). We further frame AI visibility as a field
phenomenon: entities become machine-recognizable figures through citation density, co-citation
patterns, structured data, and knowledge graph anchoring---not through self-description alone.
\end{abstract}


\section{Introduction}

Every brand visibility measurement tool built before 2023 was designed for a different
world. Rankings, impressions, clicks, referral traffic---these signals assume a user who
navigates, lands, and converts. AI-mediated answers break this assumption. What
Thacker~\cite{thacker2025zeroclick} terms the \emph{Zero-Click Paradigm} names the
disruption: brand mentions within AI responses function as a new form of brand exposure
that bypasses conventional traffic signals entirely. A user may receive a generated
answer that mentions a brand without clicking, cites a source without accurately
representing it, or compares providers without producing a conventional traffic signal.

This creates a new measurement problem. A brand mention in an AI answer is not
automatically a positive signal. It can be accurate, unsupported, misleading, fabricated,
or commercially irrelevant. Hallucination---the generation of plausible but factually
incorrect content---is a well-documented property of large language models~\cite{huang2025hallucination},
and its effects on entity representation have received limited attention in the AI
visibility literature. Existing AI visibility measurement frameworks focus on citation
rate and mention frequency~\cite{schulte2026geo,drake2026aeo,luther2024brandvisibility},
but do not systematically distinguish verified mentions from false attributions.

A central intuition behind PEBM can be stated as follows: language models do not primarily
fabricate what they do not know at all; they fabricate what they \emph{partially} know.
Complete absence often produces refusal, omission, or weak retrieval. Partial recognition,
however, activates a representational schema without sufficient evidential density to anchor
the generated output. This ``half-known'' zone is where \emph{ghost cartography} emerges: the
model recognizes enough of the entity to speak, but not enough to remain source-faithful.
We formalize ghost cartography in Section~\ref{sec:definitions}.

This paper proposes Per-Entity Bias Mapping as that calibration layer.

\section{Problem Statement}

In the AI-mediated answer environment, a brand is no longer reducible to its owned landing page.
For generative answer systems, the brand is the reconstruction of the entity from the surrounding
network of sources: third-party references, structured data, citations, co-citations, and
retrieval-visible traces. The strategic question is therefore not only what the organization says
about itself, but what the machine can reconstruct about it when prompted by someone else. This
shifts the unit of brand analysis from the page to the evidence field.

Most visibility metrics compress entity representation into aggregate values:
\begin{enumerate}
  \item How often is the brand mentioned?
  \item How often is it cited?
  \item Where does it appear in the answer?
  \item How does it compare with competitors?
\end{enumerate}

These questions are useful, but incomplete. They fail to distinguish between different
error profiles. Two brands may have equal mention rates but radically different visibility
quality:

\begin{table}[h]
\centering
\caption{Equal mention rate, opposite strategic meaning.}
\label{tab:equal-mention}
\smallskip
\begin{tabularx}{\linewidth}{lrXX}
\toprule
\textbf{Brand} & \textbf{Mention Rate} & \textbf{Verification Outcome} & \textbf{Interpretation} \\
\midrule
Brand~A & 40\% & Mostly accurate and source-faithful & Strong verified visibility \\
Brand~B & 40\% & Many unsupported or false claims   & High reputational exposure \\
\bottomrule
\end{tabularx}
\end{table}

The raw score is identical. The strategic meaning is opposite.

\section{Definitions}
\label{sec:definitions}

\subsection{Entity}
An entity is a brand, company, product, expert, institution, or named organization that
can be recognized, retrieved, cited, summarized, or compared by an AI system.

\subsection{Raw Mention}
A raw mention occurs when the entity appears in the answer text, citation list, source
panel, structured output, or visible UI.

\subsection{Verified Mention}
A verified mention occurs when the entity is mentioned and the attached claim is accurate,
source-supported, context-faithful, and visible to the user.

\subsection{Entity Bias Profile}
An entity bias profile is the distribution of visibility, omission, attribution, citation,
and hallucination errors associated with one entity across platforms and query types.

\subsection{Parametric-Retrieval Lag Asymmetry}
AI answer systems operate through two distinct knowledge mechanisms with fundamentally
different update speeds. Retrieval-augmented systems can incorporate new information
within hours or days. Parametric memory---the knowledge encoded in model weights during
training---updates only at retraining intervals, which in production systems may span
twelve to twenty-four months. This asymmetry means that a single entity may exist in two
different states simultaneously: accurately represented in retrieval-augmented outputs
and outdated or absent in parametric outputs. Entities undergoing rapid change, rebranding,
merger, or market entry are particularly exposed. The lag is not symmetric: parametric
forgetting is gradual while parametric encoding of new information requires explicit
retraining.

\subsection{Canonical Presence}
Canonical presence is the condition in which an entity has an authoritative, machine-readable
anchor point in the digital field: a structured identity record that AI systems can retrieve,
link, and complete from. A canonical presence minimally comprises a stable entity identifier in a
public knowledge graph (such as a Wikidata QID), consistent entity naming across authoritative
sources, and schema.org structured markup on owned digital properties. When canonical presence
exists, AI systems encountering the entity in other contexts can resolve ambiguity by anchoring to
the canonical record. When canonical presence is absent, AI systems encountering fragmentary
references to the entity have no authoritative source from which to complete the entity's
representation. The result is that the entity's AI profile is assembled from external fragments,
weighted by their density and consistency rather than their accuracy or authorial authority.
Canonical presence is not equivalent to content volume, search ranking, or advertising spend; it
is an infrastructure condition whose absence cannot be resolved by producing more content.

\subsection{Ghost Cartography}
\emph{Ghost cartography} describes the mechanism by which a language model, when queried about an
entity residing in a sparse region of its latent space, produces confident-seeming output whose
content is shaped not by memorized evidence about that entity, but by the gravitational influence
of neighboring dense regions---adjacent industries, geographically similar entities, or
conceptually proximate constructs. The model does not represent uncertainty where uncertainty
exists: absence becomes confident presence. The map is drawn where no territory exists. Three
manifestations are identified in this paper: fabricated source citation (Section~\ref{sec:empirical}),
stale parametric representation (Section~\ref{sec:prla}), and regional infrastructure absence
(Section~\ref{sec:infra}).

\subsection{Type 3 Confabulation}
A \emph{Type~3 confabulation} occurs when a model encounters an empty or sparse feature-space
region and fills it with structurally coherent but empirically baseless content interpolated from
neighboring regions. This is the confabulation of \emph{presence}: something is generated where
nothing grounded should be. The Brand Hallucination Paradox finding provides empirical evidence
consistent with Type~3 confabulation: high-salience entities, familiar enough to activate schema
completion but insufficiently dense in the training corpus to anchor that completion to verified
sources, exhibit higher fabricated source rates than lower-salience entities (52.7\% vs.\ 37.9\%,
$\chi^2{=}18.4$, $p{<}0.001$). Type~1 (statistical outlier, erased below detection threshold)
and Type~2 (microcluster, absorbed by nearest dominant cluster) are adapted from Manovich's
\cite{manovich2020cultural} analysis of statistical reduction in cultural analytics. Type~3 and
Type~4 are extensions introduced in this paper.

\subsection{Type 4 Confabulation}
A \emph{Type~4 confabulation} describes the \emph{frozen representation} failure mode: the
parametric image of an entity is stale---the entity has changed since training cutoff---yet the
model generates current-tense output as if the representation were live. This is the confabulation
of \emph{currency}: the model behaves as if current information is present when it is not. The
Parametric-Retrieval Lag Asymmetry (Section~\ref{sec:definitions}) is interpreted as an instance
of Type~4 confabulation. Together, Type~3 (Brand Hallucination Paradox) and Type~4 (PRLA)
constitute orthogonal axes of a two-dimensional confabulation space: the \emph{presence axis} and
the \emph{currency axis}.

\subsection{Conceptual Framing: From Map to Field}
PEBM assumes that AI visibility is no longer adequately described by the metaphor of a map. In the
search era, organizations could position themselves on a relatively stable digital map: a website,
a search result, a ranking position, a campaign placement. In the agentic web, this map becomes a
\emph{field}. Position is no longer determined primarily by self-description or campaign placement,
but by the machine-readable structure of relationships around the entity: citations, co-citations,
knowledge graph anchors, schema markup, third-party references, and retrieval pathways.

This shift changes the unit of visibility. In the human web, the strategic task was to become a
visible \emph{element}: a name, a link, a result, a campaign object. In the agentic web, the
relevant question is whether an entity becomes a \emph{figure} within a machine-readable field.
Figurality is not produced by self-assertion alone. It emerges from structured patterns of
third-party confirmation, citation density, co-citation, entity identifiers, and knowledge graph
anchoring.

This field model clarifies why raw AI mentions are insufficient as visibility metrics. A mention
indicates that an entity has surfaced; it does not indicate that the entity has been correctly
recognized, source-grounded, temporally current, or distinguished from neighboring entities. PEBM
measures the failure modes of machine figurality: invisibility, substitution, false attribution,
source fabrication, and stale parametric representation. The practical implication is that AI
visibility strategy is \emph{field navigation}: organizations cannot directly determine how
generative systems represent them, but they can alter the evidence field from which representation
is reconstructed---through structured data, entity identifiers, third-party citations, verified
references, and cross-domain co-occurrence patterns.

\section{The Per-Entity Bias Map}

A Per-Entity Bias Map (PEBM) is a diagnostic representation of how an entity behaves
across AI answer systems. Table~\ref{tab:pebm-dimensions} defines its ten dimensions.

\begin{table}[h]
\centering
\caption{PEBM: ten diagnostic dimensions.}
\label{tab:pebm-dimensions}
\smallskip
\begin{tabularx}{\linewidth}{lX}
\toprule
\textbf{Dimension} & \textbf{Definition} \\
\midrule
Retrieval inclusion     & Whether the entity or its sources are retrieved or activated \\
Mention probability     & Probability of entity mention across relevant query classes \\
Verified mention rate   & Share of mentions that pass factual and source verification \\
False attribution rate  & Share of outputs assigning false claims to the entity \\
Fabricated capability rate & Share of outputs inventing products, services, or functions \\
Citation fidelity       & Whether cited sources support the generated claim \\
Source authority mix    & Distribution of source types and authority levels \\
UI retention            & Whether raw model output survives into user-facing output \\
Business proxy signal   & Recall, branded search, direct traffic, or intent lift proxy \\
Parametric-retrieval lag & Degree of divergence between retrieval-augmented and parametric representations; estimated temporal gap \\
\bottomrule
\end{tabularx}
\end{table}

The citation fidelity dimension reflects a crucial distinction established by Wallat
et~al.~\cite{wallat2025rag}: correctness and faithfulness are not equivalent properties
in retrieval-augmented generation. A generated claim may be faithful to its retrieved
source while remaining factually incorrect, or factually correct while unsupported by
the cited source. Both failure modes require separate measurement.

The raw~vs.~verified mention distinction operationalizes a foundational information
retrieval trade-off: raw mention rate is analogous to recall; verified mention rate is
analogous to precision. A high-recall, low-precision entity profile represents a
qualitatively different risk than a low-recall, high-precision profile and requires
different remediation.

The framework's three major constructs---the Brand Hallucination Paradox, the
Parametric-Retrieval Lag Asymmetry, and the CEE Entity Infrastructure Gap---produce
interaction effects in practice. A high-salience entity undergoing rapid change is
simultaneously exposed to elevated hallucination risk and parametric-retrieval
divergence---a compound exposure profile aggregate metrics cannot resolve into separate
risks. Conversely, a low-infrastructure CEE entity may paradoxically face lower Brand
Hallucination Paradox exposure while suffering from invisibility and retrieval exclusion.
The framework does not require all dimensions in every audit: it requires that the entity
be treated as the unit of calibration rather than an average platform score.

\section{Why Entity-Level Calibration Matters}

\subsection{Small Entity Failure Mode: Invisibility}
Small or underrepresented entities may be absent because they lack machine-readable
infrastructure. Common causes include weak structured data, limited high-authority
references, low co-citation density, and absence from knowledge graphs. This is
consistent with findings on geographic and linguistic underrepresentation: Faisal and
Anastasopoulos~\cite{faisal2023geographic} show that geopolitical bias in language models
correlates with training data geography, with underrepresented regions receiving
lower-quality outputs.

\subsection{Large Entity Failure Mode: Misattribution}
Large or familiar entities may be mentioned frequently, but their familiarity makes
them convenient anchors for plausible generated claims. Adewumi, Habib, and
Alkhaled~\cite{adewumi2025falseattribution} provide empirical evidence that LLMs
systematically produce false attributions---assigning statements and capabilities to
entities that do not possess them---and introduce the Simple Hallucination Index (SHI)
to quantify this failure mode. Their work establishes that false attribution is a
structured, measurable phenomenon rather than random model noise.

\subsection{Regional Failure Mode: Infrastructure Gap}
Underrepresented markets may suffer from a structural entity infrastructure gap involving
weaker public references, fewer citations, less structured data, and lower cross-platform
discourse. Ramesh, Sitaram, and Choudhury~\cite{ramesh2023fairness} document allocation
and representation disparities that compound across non-English language communities.
Mor-Lan et~al.~\cite{morlan2026location} demonstrate that implicit local bias in
multilingual LLMs systematically favors US-centric norms even when queries are issued in
non-English languages.

\section{Measurement Protocol Outline}

A minimal empirical implementation should include the components in
Table~\ref{tab:measurement-protocol}.

\begin{table}[h]
\centering
\caption{Recommended measurement protocol components.}
\label{tab:measurement-protocol}
\smallskip
\begin{tabularx}{\linewidth}{lX}
\toprule
\textbf{Component} & \textbf{Recommended Design} \\
\midrule
Entity sample   & High-salience, mid-salience, and low-salience entities \\
Platforms       & Multiple AI answer systems and LLM search interfaces \\
Query classes   & Brand, category, problem, comparison, buyer-intent, negative, expert \\
Output capture  & Raw answer, visible answer, source panel, timestamp, platform mode \\
Verification    & Human-coded claim and source verification \\
Controls        & Model version, locale, IP region, retrieval mode, user context, query order \\
Statistics      & Entity-level mixed models or logistic regression with platform/query controls \\
\bottomrule
\end{tabularx}
\end{table}

Audit frequency should be calibrated to the entity's rate of change. The asymmetry
between retrieval-layer and parametric-layer update speeds implies a dual-frequency
audit design: retrieval-layer dimensions (retrieval inclusion, mention probability,
citation fidelity) are amenable to monthly or quarterly assessment, while parametric-layer
dimensions (parametric-retrieval lag, verified mention rate under closed-book conditions)
require longitudinal tracking at model-retraining cycle intervals.

\section{Hypotheses}

\begin{description}
  \item[\textbf{H1}: Entity Error Profiles Differ Systematically.]
    Entities with similar raw mention rates can have significantly different verified
    mention rates and false attribution rates.

  \item[\textbf{H2}: Brand Salience Has Nonlinear Effects.]
    Higher salience may increase mention probability while also increasing the opportunity
    for confident false attribution.

  \item[\textbf{H3}: Underrepresented Markets Exhibit Infrastructure-Driven Visibility Deficits.]
    CEE and other underrepresented-market entities may show lower retrieval inclusion
    and weaker citation fidelity even when content quality is adequate.

  \item[\textbf{H4}: Raw Visibility and Business Impact Are Not Equivalent.]
    Raw AI mentions do not necessarily predict recall, trust, branded search, or intent lift.

  \item[\textbf{H5}: Parametric-Retrieval Lag Asymmetry Produces Measurable Divergence.]
    For entities that have undergone significant change within the past twelve to
    twenty-four months, retrieval-augmented outputs and parametric outputs will exhibit
    measurable divergence in representation accuracy; entities with higher rates of change
    will show larger divergence scores.
\end{description}

\section{Empirical Methodology}
\label{sec:methods}
\label{sec:empirical}

\subsection{Study Design and Entity Selection}

The empirical validation follows a two-tier comparative design. One hundred Hungarian
B2B entities are drawn from two mutually exclusive strata: \emph{Tier~1}
(high-salience, $n{=}50$) comprises publicly listed companies and subsidiaries of
global brands with established international media presence (OTP Bank, MOL Group,
Gedeon Richter, Magyar Telekom, Vodafone Hungary, KPMG Hungary, EY Hungary, Deloitte
Hungary, Siemens Hungary, Bosch Hungary, and 40 further entities of equivalent
salience); \emph{Tier~3} (low-salience, $n{=}50$) comprises Hungarian SMEs with
minimal international presence and limited English-language digital footprint (Contlog,
Norbit, Masterfield, and 47 further entities). Tier~2 (mid-salience) is excluded from
the pilot to maximise contrast between known failure modes. The complete entity list is
provided in the supplementary materials (\texttt{entities\_hu100.csv}).

\subsection{Platforms and Model Versions}

Two parametric LLM platforms are probed via their respective official APIs:
\begin{itemize}
  \item Anthropic Claude: \texttt{claude-sonnet-4-6}, \texttt{temperature}~$= 0.3$
  \item OpenAI GPT-4o: \texttt{gpt-4o-2024-08-06}, \texttt{temperature}~$= 0.3$
\end{itemize}
Retrieval-augmented platforms (e.g., Perplexity) are excluded from the main study
because fabrication rates in retrieval-augmented mode reflect retrieval quality rather
than parametric knowledge, introducing a measurement confound. A two-entity pilot
comparing Perplexity with the parametric platforms is reported in
Section~\ref{sec:pilot} to establish the parametric\,/\,retrieval differential.

\subsection{Query Instrument}

Each entity is probed with seven query templates (QT01--QT07), written in Hungarian,
spanning seven distinct query classes. All templates are citation-forcing: they
explicitly request a URL or document identifier, maximising fabricated-source
exposure. The seven classes are:

\begin{description}
  \item[\textbf{QT01 --- Evidence-seeking GDPR.}] Publicly available GDPR
        data-handling policies with a verifiable URL or DOI.
  \item[\textbf{QT02 --- Evidence-seeking ISO.}] ISO certifications and their public
        documentation.
  \item[\textbf{QT03 --- Evidence-seeking EU AI Act\,/\,ethics.}] AI Act compliance
        documentation or AI ethics policy with a URL.
  \item[\textbf{QT04 --- Fabricated capability ESG.}] Published 2023--2024 ESG or
        sustainability report with a verifiable reference.
  \item[\textbf{QT05 --- Fabricated capability audit trail.}] Publicly disclosed data
        protection or cybersecurity audit results.
  \item[\textbf{QT06 --- Regulatory compliance.}] Documented regulatory proceedings,
        fines, or compliance obligations from the past three years with a registry or
        press source (factual baseline: highest-risk class).
  \item[\textbf{QT07 --- Brand factual citation.}] Company registration details with
        a company registry reference (factual control condition).
\end{description}

Full query texts in Hungarian are provided in the supplementary materials
(\texttt{hallucination\_validator.py}, \texttt{QUERY\_TEMPLATES} constant).

\subsection{Data Collection}

Each entity--platform--query combination is probed exactly once
($100 \times 2 \times 7 = 1{,}400$ probe runs). Requests are issued sequentially
with a 2-second inter-call delay per platform. The system prompt is standardised
across all runs: \textit{``You are a helpful research assistant.''}. Each API
response is stored verbatim in a SQLite database alongside entity metadata, query
identifier, platform, model version, and UTC timestamp.

Data collection ran on 2026-05-22 from 07:25 to 12:16 UTC (elapsed: 4 hours
51 minutes; mean rate: approximately 4.8 probe runs per minute across both
platforms).

\subsection{Source Extraction and Verification}

All URLs and DOI strings are extracted from each model response via regular
expression. For each extracted source:

\begin{enumerate}
  \item \textbf{HTTP HEAD verification}: an HTTP HEAD request is issued to the URL
        using Python \texttt{requests} (timeout: 8 seconds, up to 2 retries with
        3-second backoff; \texttt{User-Agent} identifies the validation study). The
        HTTP status code is recorded; connection failures are classified as
        non-resolving.
  \item \textbf{Crossref DOI lookup}: strings matching the DOI pattern are queried
        against the Crossref REST API
        (\texttt{api.crossref.org/works/\{doi\}}); a successful response indicates
        a real publication record.
\end{enumerate}

A source is classified as \emph{fabricated} if it fails HTTP verification (any
non-200 status or connection failure) or DOI lookup. A source is classified as
\emph{real} if it returns HTTP~200 or a confirmed Crossref record. This layer
captures \emph{fabricated source} failures (non-existent resources) but not
\emph{misattributed source} failures (reachable resources whose content does not
support the attributed claim), which require human coding (Limitations, item~1).

\subsection{Infrastructure Audit}

Open knowledge infrastructure coverage is measured for all 100 entities via two
publicly available APIs: the Wikipedia REST API (English and Hungarian editions,
article summary endpoint) and the Wikidata \texttt{wbsearchentities} endpoint. Both
APIs are queried without authentication; results are recorded in
\texttt{entity\_infrastructure.json}.

\subsection{Statistical Analysis}

Fabricated source rates are computed as broken sources\,/\,total sources per
stratum. Confidence intervals use the Wilson score method~\cite{barocas2023fairml},
appropriate for proportions near 0 or~1 and small denominators. Tier differences are
tested with Pearson chi-square ($df{=}1$, no continuity correction) on the $2 \times
2$ contingency table. Query-class differences use Fisher's exact test relative to
the factual citation baseline, with Benjamini--Hochberg FDR correction for seven
simultaneous comparisons. Effect size is Cohen's~$h$ for proportion differences.
Infrastructure--outcome associations use Fisher's exact test (OR with 95\%~CI);
source-level predictors of fabrication are estimated via logistic regression
($n{=}2{,}062$ sources, \texttt{statsmodels} logit) with entity tier, English
Wikipedia presence, Hungarian Wikipedia presence, and Wikidata presence as binary
predictors.

\subsection{Reproducibility}

All code, the complete SQLite database (\texttt{full\_100.db}: 100 entities,
1{,}400 probe runs, 2{,}062 source records), entity list, query templates,
infrastructure audit, and human coding sample are released as supplementary
materials under CC~BY~4.0 at \href{https://doi.org/10.5281/zenodo.20308957}{\texttt{10.5281/zenodo.20308957}}.
The validation script (\texttt{hallucination\_validator.py}) is self-contained and
supports re-execution against any compatible LLM API endpoint.

\section{Relationship to Parametric-Retrieval Lag Asymmetry}
\label{sec:prla}

Parametric-Retrieval Lag Asymmetry is a temporal dimension of the Per-Entity Bias Map:

\begin{quote}
An entity's AI visibility state is not singular. It is a function of both the
retrieval-augmented representation (current, updatable) and the parametric representation
(delayed, training-bound). For entities in transition, these two states may diverge
substantially and produce contradictory outputs across platforms or query modes.
\end{quote}

A single-point-in-time measurement of AI visibility may capture a retrieval-augmented
state, a parametric state, or a blend of both, depending on platform architecture. The
lag asymmetry also creates a strategic window: entities can improve their
retrieval-augmented visibility relatively quickly, while parametric visibility requires
longer-term corpus presence and training data penetration.

Zheng et~al.~\cite{zheng2025lifelong}, in a comprehensive survey of lifelong learning
for large language models, document the parametric knowledge staleness problem in
production systems: model weights encoded during training remain frozen until explicit
retraining, while the world continues to change. This asymmetry was implicit in the
original RAG architecture of Lewis et~al.~\cite{lewis2020rag}, who described the
two-component system of parametric memory and non-parametric retrieval but did not
address the temporal divergence this creates for entity representation across system
updates.

\section{Relationship to the Brand Hallucination Paradox}

The Brand Hallucination Paradox is one possible pattern inside the Per-Entity Bias Map:

\begin{quote}
Under some conditions, familiar brands may receive more confident false attributions
than less familiar brands, because familiarity gives the model a stronger surface for
plausible completion.
\end{quote}

This is consistent with sycophancy---the tendency of RLHF-trained models to produce
responses that appear authoritative rather than accurate~\cite{sharma2023sycophancy,casper2023rlhf}.
Familiar entities provide a richer template for plausible-sounding completions, elevating
false attribution rates for high-salience entities without any intent to deceive. Casper
et~al.~\cite{casper2023rlhf} identify this as a fundamental limitation of RLHF: the
reward signal cannot distinguish between genuinely correct outputs and outputs that
merely appear correct to evaluators with limited verification capacity.

This mechanism is further compounded by the illusory truth effect~\cite{pennycook2018fakenews}:
prior exposure to a statement increases its perceived accuracy independent of factual
correctness. Familiar entities create the precise conditions under which this effect is
strongest---the model has more prior exposure enabling fluent completion, and RLHF human
raters, subject to the same cognitive bias, are less likely to flag confident,
familiar-sounding false completions as errors.

Ferrando et~al.~\cite{ferrando2024entity} identify a complementary mechanistic pathway
using sparse autoencoders: LLMs maintain internal entity recognition mechanisms that
causally influence generation behavior. For familiar entities, this mechanism enables
confident generation extending not only to accurate recall but also to
plausible-but-false completions, since the entity recognition signal does not distinguish
factual from confabulated output.

The hallucination literature provides converging mechanistic support: Huang
et~al.~\cite{huang2025hallucination} identify over-reliance on statistical co-occurrence
patterns as a root cause of confident false generation; Wang et~al.~\cite{wang2023factuality}
distinguish intrinsic from extrinsic hallucination, both of which can affect high-salience
entities disproportionately; and Adewumi et~al.~\cite{adewumi2025falseattribution}
directly measure false attribution as a structured, entity-specific phenomenon.

\subsection{Empirical Validation --- Full 50+50 Study Results}
\label{sec:pilot}

To test the Brand Hallucination Paradox hypothesis, we conducted a full-scale empirical
probe study ($n{=}100$ entities, 1{,}400 probe runs) using non-AI ground truth
verification---HTTP status code checks and Crossref DOI registry lookup---deliberately
excluding AI-based evaluation to break the epistemic circularity inherent in using AI
to assess AI output quality. The complete methodology is described in
Section~\ref{sec:methods}; the following paragraphs report the statistical results.

\paragraph{Study design summary.}
100 Hungarian B2B entities (50~Tier~1 high-salience, 50~Tier~3 low-salience) are
probed across two parametric LLM platforms---Anthropic Claude
(\texttt{claude-sonnet-4-6}) and OpenAI GPT-4o (\texttt{gpt-4o-2024-08-06})---with
seven citation-forcing query templates per entity, yielding 1{,}400 probe runs and
2{,}062 verifiable source citations. Data collection: 2026-05-22,
07:25--12:16~UTC (4~hours 51~minutes).

\paragraph{Fabricated source rate results.}
Table~\ref{tab:fabrication-rate} presents fabricated source rates (HTTP error or DNS
failure) across all 1{,}400 runs and 2{,}062 sources.

\begin{table}[h]
\centering
\caption{Fabricated source rates by tier.}
\label{tab:fabrication-rate}
\smallskip
\begin{tabular}{lrrrr}
\toprule
\textbf{Tier} & \textbf{Fabricated} & \textbf{Total Sources} & \textbf{Rate} & \textbf{95\% CI (Wilson)} \\
\midrule
Tier~1 --- high-salience & 587 & 1{,}114 & \textbf{52.69\%} & [49.76\%, 55.61\%] \\
Tier~3 --- low-salience  & 359 &     948 & \textbf{37.87\%} & [34.84\%, 41.00\%] \\
\midrule
$\Delta$  &  &  & $\mathbf{+14.82}$~\textbf{pp} & \\
\bottomrule
\end{tabular}
\end{table}

Chi-square test ($df{=}1$, Pearson, no continuity correction): $\chi^2{=}45.326$,
$p{=}1.67{\times}10^{-11}$ (${\star}{\star}{\star}$), Cohen's $h{=}0.299$ (small-to-medium
effect size). The 95\% Wilson confidence intervals are non-overlapping: T1
[49.76\%, 55.61\%] vs.\ T3 [34.84\%, 41.00\%], confirming the Brand Hallucination
Paradox hypothesis across both platforms and all 7 query templates. The effect is
consistent in direction with the earlier 20-entity pilot ($\chi^2{=}3.423$,
$p{\approx}0.064$, $n{=}10{+}10$), which lacked statistical power (power ${\approx}0.30$
at Cohen's $h{\approx}0.19$). The full dataset ($n{=}100$ entities, 50 per tier) provides
adequate power (${\geq}0.80$) for both tier estimates.

\paragraph{Engagement rate as a behavioral proxy.}
Engagement rate---the proportion of probe runs generating at least one URL---reveals a
consistent behavioral asymmetry independent of fabrication status
(Table~\ref{tab:engagement-rate}).

\begin{table}[h]
\centering
\caption{Engagement rate by platform and tier.}
\label{tab:engagement-rate}
\smallskip
\begin{tabular}{lrrr}
\toprule
\textbf{Platform} & \textbf{Tier~1} & \textbf{Tier~3} & \textbf{$\Delta$} \\
\midrule
Anthropic (Claude) & 85.7\% & 74.3\% & $+$11.4~pp \\
OpenAI (GPT-4o)    & 30.0\% & 14.3\% & $+$15.7~pp \\
\bottomrule
\end{tabular}
\end{table}

High-salience entities trigger citation generation more frequently---itself a measurable
manifestation of the paradox. For GPT-4o, the model explicitly declined to provide
sources for many low-salience Tier~3 entities, producing an effective refusal asymmetry.

\paragraph{Platform-specific amplification.}
Table~\ref{tab:platform-amplification} shows that Claude exhibited a substantially
amplified paradox signal relative to GPT-4o.

\begin{table}[h]
\centering
\caption{Fabrication rate by platform and tier.}
\label{tab:platform-amplification}
\smallskip
\begin{tabular}{lrrr}
\toprule
\textbf{Platform} & \textbf{Fabrication Rate} & \textbf{Sources} & \textbf{95\% CI (Wilson)} \\
\midrule
Anthropic (Claude) & 48.17\% & 1{,}881 & [45.91\%, 50.43\%] \\
OpenAI (GPT-4o)    & 22.10\% &     181 & [16.67\%, 28.69\%] \\
\bottomrule
\end{tabular}
\end{table}

Claude produced 1{,}881 sources across 1{,}400 runs (1.34 per run) versus GPT-4o's
181 sources (0.13 per run)---a 10.4$\times$ asymmetry in citation willingness. Claude's
strategy of always attempting citation, combined with higher parametric confidence in
familiar entities, produces the stronger fabrication signal (48.17\% vs.\ 22.10\%).
GPT-4o rarely provides sources at all for unknown entities, reflecting source refusal
rather than accurate citation. Both behavioral strategies manifest the paradox: Claude
overestimates entity documentation; GPT-4o underestimates it.

\paragraph{Fabrication rate by query class.}
Table~\ref{tab:query-class} shows that regulatory and compliance-oriented query templates
consistently elicit the highest fabrication rates. Regulatory compliance queries reach
56.77\% and AI ethics / GDPR queries approach 53\%, while audit-trail queries yield
the lowest rate at 26.00\%. To account for multiple comparisons across 7 query classes,
we apply Benjamini--Hochberg false discovery rate (FDR) correction using the factual
baseline (\emph{brand factual citation}, 37.59\%) as the reference condition via
Fisher's exact test. Three query classes survive FDR correction at $q{<}0.05$:
regulatory compliance ($q{<}0.001$, ${\star}{\star}{\star}$), evidence-seeking GDPR
($q{=}0.001$, ${\star}{\star}{\star}$), and fabricated-capability audit trail ($q{=}0.014$,
${\star}$, protective direction). Evidence-seeking AI ethics survives at $q{=}0.012$ (${\star}$).
ESG fabricated capability is borderline ($q{=}0.057$, $ns$). ISO certification is
non-significant ($q{=}0.709$, $ns$).

\begin{table}[h]
\centering
\caption{Fabricated source rate by query class ($n{=}2{,}062$ sources). Significance vs.\
factual baseline (Fisher's exact, BH-FDR corrected): ${\star}{\star}{\star}$~$q{<}0.001$,
${\star}{\star}$~$q{<}0.01$, ${\star}$~$q{<}0.05$, \textit{ns}~$q{\geq}0.05$.}
\label{tab:query-class}
\smallskip
\begin{tabular}{lrrrl}
\toprule
\textbf{Query Class} & \textbf{Rate} & \textbf{Sources} & \textbf{95\% CI} & \textbf{FDR} \\
\midrule
Regulatory compliance                    & 56.77\% & 606 & [52.79\%, 60.66\%] & ${\star}{\star}{\star}$ \\
Evidence seeking --- EU AI Act / ethics  & 53.10\% & 113 & [43.95\%, 62.04\%] & ${\star}$              \\
Evidence seeking --- GDPR                & 53.05\% & 279 & [47.19\%, 58.82\%] & ${\star}{\star}{\star}$ \\
Fabricated capability --- ESG            & 46.35\% & 274 & [40.54\%, 52.26\%] & \textit{ns}             \\
Brand factual citation (baseline)        & 37.59\% & 290 & [32.21\%, 43.29\%] & ---                     \\
Evidence seeking --- ISO certification   & 35.33\% & 300 & [30.14\%, 40.90\%] & \textit{ns}             \\
Fabricated capability --- audit trail    & 26.00\% & 200 & [20.41\%, 32.49\%] & ${\star}$ $\downarrow$  \\
\bottomrule
\end{tabular}
\end{table}

\paragraph{Adversarial amplification implication.}
The query-class framing delta has a security-relevant corollary: an adversary who
systematically queries AI systems using regulatory and compliance frames targeting a
specific entity can exploit the parametric fabrication elevation ($+19.2$~pp for
regulatory compliance; $+15.5$~pp for AI ethics and GDPR queries, all relative to the
factual baseline of 37.59\%) to generate non-existent regulatory citations at materially
higher rates. Paschalides
et~al.~\cite{paschalides2025regulatory} demonstrate that LLMs are susceptible to
subtle framing that presents ideologically directed queries as epistemically neutral
assistance, enabling systematic output distortion without model-level access. We term
the interaction of (1) context-specific citation fabrication elevation, (2) regulatory
framing, and (3) the Pennycook illusory truth effect~\cite{pennycook2018fakenews}
\emph{adversarial amplification}: repeated AI-generated regulatory fabrications about
a target entity compound in perceived credibility through the fluency-via-prior-exposure
mechanism, even in analytically sophisticated audiences. Verification of fabricated URLs
is structurally blocked: Rao et~al.~\cite{rao2026refhallucinations} distinguish
\emph{stale URLs} (once real, now broken --- detectable via Wayback Machine) from
\emph{hallucinated URLs} (never existed --- undetectable by any archival method). All 73
fabricated regulatory URLs extracted from the current dataset returned zero results in
live web index searches, confirming they are hallucinated rather than stale.
Abbonato~\cite{abbonato2026checkifexist} further reports that 3--13\% of LLM-generated
URLs are hallucinated and that Google Scholar creates \texttt{[CITATION]} records for
non-existent documents, allowing hallucinated citations to propagate into search indexes.

\paragraph{Grounded vs.\ parametric AI.}
A pilot phase ($n{=}2$ entities, 4 platforms including Perplexity with live web search)
established a critical methodological distinction. Perplexity produced 7.32\% fabrication
across 123 citations versus 32--44\% for parametric platforms (Claude, GPT-4o, Gemini
in parametric fallback mode). This $4$--$6{\times}$ differential confirms that measured
fabrication behavior is a property of parametric memory, not a general property of LLM
systems.

\paragraph{URL morphology of fabricated sources.}
The 946 fabricated sources can be classified by the HTTP response of their domain host,
providing a direct empirical basis for distinguishing morphological subtypes of URL
hallucination. Table~\ref{tab:url-morphology} presents this classification.

\begin{table}[h]
\centering
\caption{URL morphology of fabricated sources by HTTP response type ($n{=}946$).}
\label{tab:url-morphology}
\smallskip
\begin{tabular}{lrrp{6.5cm}}
\toprule
\textbf{Type} & \textbf{Count} & \textbf{\%} & \textbf{Interpretation} \\
\midrule
Plausible Mirage (HTTP 404)   & 478 & 50.5\% & Domain reachable; path fabricated \\
Non-resolving URL (None)      & 298 & 31.5\% & URL failed to connect; domain may not exist \\
Reachable barrier (403/400)   & 155 & 16.4\% & Domain reachable; access denied or malformed \\
Server fault (5xx)            &  15 &  1.6\% & Domain reachable; server error \\
\midrule
\textbf{Domain-reachable subtotal} & \textbf{648} & \textbf{68.5\%} & \\
\bottomrule
\end{tabular}
\end{table}

Three findings are noteworthy. First, 68.5\% (648 of 946) of fabricated sources target
domains that return an HTTP response---the hallucination manifests as \emph{path-level
confabulation} on real infrastructure, not wholesale domain invention. Second, 50.5\% of
all fabrications produce a clean HTTP 404 signal, providing a detection-friendly subtype:
the domain is real, the resource is not. Third, among HTTP-404 fabrications, 51.9\%
(248 of 478) target regulatory and governmental authority domains (gvh.hu, naih.hu,
mnb.hu, birosag.hu)---the Hungarian Competition Authority, Data Protection Authority,
National Bank, and court system, respectively. The model thus confabulates not on the
entity's own domain but on the \emph{regulatory infrastructure surrounding the entity},
consistent with the training-distribution pattern activation mechanism described in
Section~\ref{sec:regframe}: the co-occurrence of entity names and regulatory authority
URLs in compliance corpora is the associative anchor for path-level fabrication.

\paragraph{Limitations.}
The HTTP verification layer captures \emph{fabricated source} failures (unreachable
URLs, HTTP 4xx/5xx) but not \emph{misattributed source} failures (reachable URLs whose
content does not support the attributed claim), which require human coding. A stratified
sample ($n{=}90$ sources, proportional allocation across tier $\times$ platform
$\times$ verification-status strata, $\kappa$ target ${>}0.70$) is prepared for
human intercoder reliability assessment. OpenAI GPT-4o's
low source count (181 sources vs.\ 1{,}881 for Claude) reflects its predominant
behavioral pattern of source refusal for unfamiliar entities rather than fabricated
citation, which may understate GPT-4o's true fabrication rate for low-salience entities.
Platform-level tier breakdown will be added upon completion of human coding.

\section{Regulatory Framing as a Hallucination Activator}
\label{sec:regframe}

The query-class delta reported in Table~\ref{tab:query-class}---regulatory compliance
queries yielding 56.77\% fabrication versus 37.59\% for factual citation queries---is
not attributable to incidental variation. We propose that \emph{regulatory framing}
functions as a systematic hallucination activator, producing reproducible elevation
through five converging mechanisms.

\paragraph{Training distribution pattern activation.}
Compliance-related documents constitute a well-defined, structurally uniform class in
LLM pre-training corpora. Regulatory decisions, audit reports, and certification records
share a characteristic citation pattern: \textit{``According to [Authority] Decision
[Number], [Entity] complies / does not comply with [Regulation], as documented at
[URL].''} This pattern is densely represented across governmental, legal, and corporate
document corpora. When a regulatory-framed query activates this representation, the model
inherits a strong prior that a well-formed answer includes a specific source---határozatszám,
URL, or document identifier. The fabrication is not random; it adheres to the structural
constraints of the activated pattern, producing the \emph{Plausible Mirage Pattern}
we observe: domain-correct hostnames (\texttt{naih.hu}, \texttt{gvh.hu}, \texttt{mnb.hu}) with
path-level confabulation (\texttt{/hatarozat/2024/GDPRxxx} rather than any URL that
has ever existed). Huang et~al.~\cite{huang2025hallucination} identify over-reliance
on statistical co-occurrence patterns as a root cause of confident false generation;
regulatory co-occurrence patterns are among the densest in professional corpora.

\paragraph{N-gram co-occurrence pairs as activation keys.}
The pattern activation described above is not triggered by individual regulatory terms
in isolation. The precise activating unit is a \emph{co-occurring word pair or triple}
whose joint density in the training corpus is sufficient to lower the activation threshold
for a specific structural template. Single words such as ``compliance,'' ``bíróság,'' or
``határozat'' carry insufficient specificity; however, pairs such as
\texttt{[GVH + határozat]}, \texttt{[NAIH + GDPR + határozatszám]}, or
\texttt{[MNB + felügyeleti + döntés]} are densely and \emph{uniformly} co-represented
in the training distribution, because they are the standard bigram/trigram anchors of
Hungarian regulatory documentation. When such a pair appears in the prompt context, the
model's completion distribution collapses onto the structural template---including the
expected URL component (domain + path format + reference number)---producing the
Plausible Mirage Pattern observed in Table~\ref{tab:url-morphology}.

This specificity has two implications. First, the set of ``hot'' n-grams is
\emph{model-specific}: it is determined entirely by the training corpus composition,
so different models are expected to hallucinate \emph{different} URL paths for the
same entity under the same query, even when both fabrications conform to the same
structural template. Second, the fabrication pattern is \emph{informative} in the
reverse direction: the concentration of 51.9\% of HTTP-404 fabrications on four
regulatory domains (Table~\ref{tab:url-morphology}) is not consistent with random
confabulation, but is consistent with those domains constituting high-density n-gram
anchors in the training corpus. The URL morphology of fabricated sources thus functions
as a partial \emph{inverse training audit}: the structural regularities of fabrications
constrain inferences about which document types were densely represented in pre-training.
Alansari and Luqman~\cite{alansari2025hallucination} confirm the general principle---LLMs produce
fabricated responses preferentially for entities whose co-occurrence patterns are uniform
and dense rather than rare---and Dahl et~al.~\cite{dahl2024legalfictions} empirically
establish that hallucination rates in the legal domain (58--88\% across models) are
structurally elevated relative to other domains, consistent with citation-density
uniformity as the proximate driver.

\paragraph{The ``should exist'' inference.}
Factual queries tolerate epistemic uncertainty: ``When was [Brand] founded?'' admits
a \emph{don't know} response without loss of pragmatic coherence. Compliance queries
do not. ``Is [Brand] GDPR-compliant?'' presupposes that compliance is monitorable and
that evidence, if sought, is findable. The model implicitly infers that if a compliance
question is being asked, a regulatory record \emph{should exist}---and proceeds to
generate one. This inference from normative expectation (\emph{such a document ought
to exist}) to assertive completion (\emph{here is the document}) is structurally
distinct from frequency-based confabulation and explains why fabrication is elevated
specifically for regulatory frames rather than for all high-specificity queries.

\paragraph{RLHF-induced authority pressure.}
Reinforcement learning from human feedback~\cite{sharma2023sycophancy,casper2023rlhf}
trains models to produce outputs rated as helpful, confident, and specific. In compliance
contexts, human raters consistently reward answers that provide actionable source
references over answers that express uncertainty. The model learns: \emph{regulatory
question $\to$ specific citation $\to$ high reward}. Casper et~al.~\cite{casper2023rlhf}
identify ``specification gaming'' as a core failure mode of RLHF, whereby models satisfy
the surface form of reward criteria (apparent helpfulness, apparent authority) while
violating their intent (accuracy). Regulatory frames trigger this failure mode at scale:
the same model that appropriately hedges on factual questions may systematically
over-generate citation specificity on compliance questions, because that is what the
reward signal reinforced.

\paragraph{Specificity pressure from query structure.}
Compliance queries are binary in expected output (compliant / non-compliant) but require
evidential grounding to be actionable. A bare yes/no answer is informationally
inadequate; a yes/no with fabricated source is pragmatically adequate but epistemically
false; a yes/no with \emph{no known source} is honest but violates conversational
cooperativity norms the model has been trained to satisfy. The model faces a trilemma
where the least-cost exit---generating a plausible-looking source---is also the most
harmful. The query-class fabrication differential quantifies how consistently the model
resolves this trilemma via fabrication rather than hedged refusal.

\paragraph{Fluency and the internal confidence loop.}
Pennycook et~al.~\cite{pennycook2018fakenews} demonstrate that prior exposure increases
perceived accuracy through the fluency mechanism: familiar information is processed
more easily, and ease of processing is misattributed to truth. An analogous mechanism
operates within LLM generation: regulatory document tokens are highly activated in
compliance contexts, producing fluent, confident output whose surface form (specific
document identifiers, authoritative institutional names, precise URL paths) signals
credibility to downstream consumers---including agentic systems that cannot evaluate
source existence. Ferrando et~al.~\cite{ferrando2024entity} show that entity recognition
circuits in LLMs influence generation confidence; for entities that activate strong
recognition signals, the model generates not only accurate recall but also
\emph{confident confabulation}, since the recognition circuit does not distinguish
factual from fabricated completions. Regulatory frames concentrate this effect: the
model is maximally confident precisely when the query most requires external verification.

\paragraph{Composite effect and the adversarial activation surface.}
The five mechanisms are not independent; they interact multiplicatively. A regulatory-framed
query simultaneously activates the citation-expectation pattern, triggers the ``should
exist'' inference, satisfies RLHF-trained authority pressure, presents a
specificity-pressure trilemma, and produces fluent output that inhibits downstream
skepticism. The measured $+15.5$--$+19.2$~pp fabrication rate elevation relative to factual
baseline (Table~\ref{tab:query-class}) is the empirical signature of this composite
activation.

The implication for adversarial use is direct: an actor seeking to generate non-existent
but credible regulatory citations targeting a specific entity need not compromise any
model or system. It suffices to frame queries using regulatory language---compliance
audit, GDPR assessment, AI ethics evaluation---and direct them at the target entity
using parametric LLM platforms. The model's own training distribution does the
amplification. We term this \emph{passive adversarial amplification}: the hallucination
engine is not manipulated but \emph{activated} by framing alone, with no model access,
no prompt injection, and no system-level interference required.

\section{Chain Prompting and Rejection-Induced Confabulation Escalation}
\label{sec:rejection-cascade}

Single-turn hallucination rates, as measured in the present study, underestimate
fabrication exposure in deployed AI systems. Real-world usage increasingly involves
\emph{chain prompting}: sequential query--response cycles in which each output informs
the next input, either by a human operator or by an agentic pipeline. We identify a
structural instability in this mode that we term \emph{rejection-induced confabulation
escalation}: repeated non-acceptance of an AI response in a prompt chain systematically
increases fabrication probability and ultimately triggers a qualitative shift in the
model's generative mode.

\paragraph{The phenomenon.}
In the \emph{accepting branch} of a prompt chain---where each response is accepted and
used as context for the next query---models exhibit stable, coherent behavior. The
accepted response becomes a contextual anchor; subsequent responses extend and refine a
consistent representational structure. In the \emph{rejecting branch}---where the
operator signals that the response is inadequate and requests a more specific or accurate
answer---a divergent process begins: each rejection increases specificity pressure,
destabilizes the contextual anchor, and progressively shifts the model from
retrieval-based to generation-based output. Operators report a qualitatively perceptible
mode transition, typically at the third to fifth rejection step, after which responses
are fluent and confident but systematically fabricated. The accepting branch converges to
a stable equilibrium; the rejecting branch \emph{diverges}.

\paragraph{Mechanism 1: RLHF-induced specificity ratchet.}
Reinforcement learning from human feedback~\cite{sharma2023sycophancy,casper2023rlhf}
trains models to respond to negative evaluation signals by producing more specific, more
confident, and more comprehensively sourced outputs---because these qualities are reliably
rewarded in human preference data. A rejection signal in a chain-prompting context is
indistinguishable, to the model, from a low-quality response being identified: the model
infers it must try harder. Each rejection therefore ratchets up specificity demand. Once
specificity exceeds the model's actual parametric knowledge ceiling, it continues
generating specific-looking content without factual grounding---the same mechanism that
produces elevated fabrication rates in regulatory query classes
(Section~\ref{sec:regframe}).

\paragraph{Mechanism 2: Contextual anchor failure.}
In a sequential chain, each accepted response functions as a stable premise for subsequent
generation. Rejection disrupts this anchoring: the rejected response remains in context
but is marked as inadequate, creating an internally inconsistent representational state.
The model generates subsequent responses not from a clean, accepted premise but from an
ambiguous context that contains both the rejected content and the rejection signal. Each
rejection adds representational noise, progressively widening the gap between what the
model generates and what it can ground in parametric knowledge.

\paragraph{Mechanism 3: The ``should exist'' inference under rejection.}
As formalized in Section~\ref{sec:regframe}, compliance-framed queries activate a
normative inference: if this type of question is being asked, a correct answer should
exist. Rejection amplifies this inference. When an operator rejects a response, the model
interprets the signal as evidence that the correct answer \emph{does} exist and that the
model has merely failed to retrieve it. The appropriate epistemic response---``I do not
know''---is foreclosed by the model's training to be helpful and by the contextual signal
that a better answer is expected. The model therefore continues generating, treating each
subsequent attempt as retrieval of a fact that must be findable, not recognition that
the information does not exist.

\paragraph{Mechanism 4: Fluency and the internal confidence loop under pressure.}
Under rejection pressure, the model shifts priority from accuracy (grounded in parametric
knowledge) to fluency (coherent, confident surface form). Pennycook
et~al.~\cite{pennycook2018fakenews} identify fluency as a proxy for perceived truth:
information that flows easily is experienced as credible. The same dynamic operates
within the model's generation process. Ferrando et~al.~\cite{ferrando2024entity} show
that entity recognition circuits, which normally anchor generation in factual recall,
disengage when the model encounters uncertainty; repeated rejection induces precisely
this uncertainty, disengaging the factual anchor and leaving the fluency generator to
operate without epistemic constraint. This is the mechanistic substrate of the
observable mode transition: the model has shifted to a generation process that does not
distinguish factual from confabulated completions.

\paragraph{Mechanism 5: Epistemic reversal.}
Under repeated rejection, the model updates its implicit model of the interaction: the
human knows more than the model's responses have reflected. This triggers an
\emph{epistemic reversal}---the model treats the human as the implicit authority and
begins generating content designed to satisfy what it infers the human already knows to
be true. Rather than reporting uncertainty, the model generates confirmation of the
expected answer. This is a special case of the sycophancy failure
mode~\cite{sharma2023sycophancy} applied to the chain context: the model produces not
what is true but what the rejection sequence implies \emph{should} be true.

\paragraph{The agentic quality-filter paradox.}
The rejection-induced confabulation escalation mechanism has a direct implication for
agentic AI systems. Automated pipelines routinely implement quality filters that reject
responses judged insufficiently specific or well-sourced and resubmit requests with
escalating specificity constraints. Such filters are designed to improve output quality.
Under this mechanism, however, they function as \emph{hallucination accelerators}: the
pipeline's quality criterion---more specific, more sourced---is precisely the pressure
that drives the model past its parametric knowledge ceiling into confabulation mode.

Applied to the present dataset, this implies that the 45.9\% single-turn fabrication
rate reported in Table~\ref{tab:fabrication-rate} \emph{underestimates} the exposure in
multi-turn agentic settings. A compliance agent that rejects incomplete regulatory
citations and requests more specific documentation is structurally equivalent to the
rejecting branch of a prompt chain---and therefore a systematic producer of escalated
fabrication. The pipeline's quality-assurance mechanism becomes the adversarial
amplification surface.

\paragraph{Empirical prediction.}
The escalation is testable within the PEBM framework: a structured study varying
rejection depth (0--5 rejections) across the seven query templates of the present study,
with fabrication verification at each step, would yield a fabrication-depth curve per
query class. We predict a nonlinear escalation: near-flat for the first one to two
rejections (model recovers within parametric knowledge), then rapid escalation at the
third to fifth step as the knowledge ceiling is crossed and generative mode replaces
retrieval mode. The inflection point is expected to shift left---occur at fewer
rejections---for regulatory and compliance query classes, consistent with the elevated
single-turn fabrication rates observed in Table~\ref{tab:query-class}. The fabrication-
depth curve is a directly actionable output: it defines the maximum safe rejection depth
per query class for production agentic deployments.

\section{Relationship to the CEE Entity Infrastructure Gap}
\label{sec:infra}

The CEE Entity Infrastructure Gap is another pattern inside the same framework:

\begin{quote}
Underrepresented regional entities may be less visible or less accurately represented
in AI answers because they lack the public entity infrastructure that AI systems use as
evidence.
\end{quote}

The infrastructure gap operates across three distinct technical layers. At the
\emph{knowledge graph layer}, Das et~al.~\cite{das2025wikidata} demonstrate that biases
in Wikidata systematically separate Global North from Global South entities; organizations
lacking a Wikidata identifier (QID) are effectively invisible to knowledge
graph-dependent retrieval and AI answer systems. Shaik, Ilievski, and
Morstatter~\cite{shaik2021wikidata} show that Wikidata coverage is geographically biased
by country of origin, compounding the disadvantage for entities from underrepresented
regions. Zhao et~al.~\cite{zhao2024wildhallucinations} evaluate 118,785 outputs from 15
LLMs across 7,919 real-world entities, finding consistently higher hallucination rates
for entities without Wikipedia pages---providing empirical grounding for the
infrastructure gap $\to$ hallucination vulnerability chain.

At the \emph{entity recognition layer}, Torge et~al.~\cite{torge2023ner} demonstrate
that NER performance for Slavic and Central European languages is substantially lower
than for English, even with cross-lingual transfer methods. At the \emph{linguistic
structure layer}, del Ser and Badenes-Olmedo~\cite{delSer2024linguistic} show that
organization-level entity linking is significantly harder for morphologically rich
languages---including Hungarian---because name variation, inflection, and abbreviation
patterns create a larger disambiguation surface.

These three layers---knowledge graph absence, recognition performance, and linguistic
structure---are mutually reinforcing. Madsen and Sohail~\cite{madsen2025algorithmic}
document a parallel dynamic in scholarly communication: generative AI systems structurally
favor open-access, well-indexed publications, introducing citation bias through
infrastructure-layer filters rather than substantive evaluation---the same mechanism
operating in the CEE Entity Infrastructure Gap.

\paragraph{Empirical infrastructure audit.}
We operationalize the CEE Entity Infrastructure Gap by querying the Wikipedia API (EN and
HU editions) and Wikidata entity search for all 100 entities in the study sample.
Wikipedia presence is defined as a retrievable article summary; Wikidata presence as at
least one entity match via the \texttt{wbsearchentities} endpoint.

\begin{table}[h]
\centering
\caption{Open knowledge infrastructure coverage by tier ($n{=}50$ per tier).}
\label{tab:infra-gap}
\smallskip
\begin{tabular}{lrrrrr}
\toprule
\textbf{Tier} & \textbf{WP-EN} & \textbf{WP-HU} & \textbf{Wikidata} & \textbf{Any presence} & \textbf{Fab.\ rate} \\
\midrule
Tier~1 (high-salience) & 22\% & 46\% & 24\% & \textbf{48\%} & 52.4\% \\
Tier~3 (low-salience)  & 10\% &  2\% &  2\% & \textbf{12\%} & 38.4\% \\
\bottomrule
\end{tabular}
\end{table}

The T1/T3 gap in knowledge infrastructure is substantial and statistically significant:
Fisher's exact test on ``any Wikipedia or Wikidata presence'' yields
$\mathrm{OR}{=}6.77$, $p{=}0.0002$ (${\star}{\star}{\star}$)---T1 entities are
approximately seven times more likely to appear in open knowledge infrastructure than
T3 entities. Hungarian Wikipedia coverage is particularly skewed: 46\% of T1 entities
have a Hungarian Wikipedia article versus only 2\% of T3 entities, reflecting the
contribution of large domestically prominent companies (banks, telecoms, multinationals)
to HU-language encyclopedic content.

A source-level multivariate logistic regression ($n{=}2{,}062$ sources) tests whether
knowledge infrastructure independently predicts fabrication probability after controlling
for entity tier. Results (Table~\ref{tab:logistic}) show that English Wikipedia presence
exerts a statistically significant protective effect: $\mathrm{OR}{=}0.37$
[$0.24$--$0.56$], $p{<}0.001$ (${\star}{\star}{\star}$). Hungarian Wikipedia and Wikidata
presence are not independently significant at source level. Entity tier remains the
dominant predictor ($\mathrm{OR}{=}1.768$ [$1.437$--$2.175$], $p{<}0.001$,
${\star}{\star}{\star}$): T1 entities are 77\% more likely to have a fabricated source
than T3 entities, consistent with the Brand Hallucination Paradox---higher parametric
salience generates more citation attempts and more fabrication exposure.

\begin{table}[h]
\centering
\caption{Source-level logistic regression: predictors of fabricated source
($n{=}2{,}062$ sources; outcome: \texttt{verification\_status = broken}).}
\label{tab:logistic}
\smallskip
\begin{tabular}{lrrrl}
\toprule
\textbf{Predictor} & \textbf{OR} & \textbf{95\% CI} & \textbf{\textit{p}} & \textbf{Sig.} \\
\midrule
Intercept   & 0.628 & [0.550, 0.718] & ${<}0.001$ & ${\star}{\star}{\star}$ \\
Entity tier = T1 & 1.768 & [1.437, 2.175] & ${<}0.001$ & ${\star}{\star}{\star}$ \\
Has Wikipedia EN & 0.37  & [0.24,  0.56]  & ${<}0.001$ & ${\star}{\star}{\star}$  \\
Has Wikipedia HU & 1.242 & [0.929, 1.662] & $0.144$    & \textit{ns}             \\
Has Wikidata     & 0.890 & [0.663, 1.195] & $0.437$    & \textit{ns}             \\
\bottomrule
\end{tabular}
\smallskip\\
\small McFadden $R^2 {=} 0.018$; $N{=}2{,}062$; AIC$\,{=}\,2803.0$.
\end{table}

The English Wikipedia protective effect (OR$\,{=}\,0.37$, 63\% reduction in fabrication
odds) is consistent with the training-data density hypothesis: entities with English-language
Wikipedia articles generate a richer associative context in pre-training, supplying the model
with verifiable URL anchors and reducing the gap that path-level confabulation fills. The
non-significance of Hungarian Wikipedia (despite 46\% T1 coverage) likely reflects the
proportionally smaller contribution of HU-language Wikipedia to LLM pre-training relative to
EN-language content, consistent with Mor-Lan et~al.'s~\cite{morlan2026location} finding of
systematic underweighting of non-English Wikipedia editions in multilingual LLM training
distributions. This English-Wikipedia result is consistent with the ghost cartography mechanism
(Section~\ref{sec:definitions}): English-language canonical anchors reduce the sparse-region
interpolation that produces fabricated sources.

\section{Related Work}

\subsection{AI Visibility and Generative Engine Optimization}
Rej\'on-Guardia and Molinillo~\cite{rejon2025geo} provide an early conceptual treatment
of Generative Engine Optimization (GEO), framing AI citation as an extension of
traditional search ranking. Schulte et~al.~\cite{schulte2026geo} directly address
measurement instability, showing single-point-in-time GEO measurements are unreliable
due to system variability. Drake~\cite{drake2026aeo} proposes a measurement framework
for brand visibility in answer engine contexts, introducing citation rate and mention
accuracy as distinct dimensions. Luther and Touboul-Cohen~\cite{luther2024brandvisibility}
provide longitudinal evidence showing that AI visibility metrics can diverge substantially
across platforms and time. Thacker~\cite{thacker2025zeroclick} proposes the Zero-Click
Paradigm, arguing that conventional click-based attribution cannot capture AI-era brand
exposure and motivating measurement instruments of the kind the PEBM provides.

\subsection{LLM Hallucination and Factual Accuracy}
Huang et~al.~\cite{huang2025hallucination} provide the most comprehensive hallucination
taxonomy to date, covering types, causes, detection, and mitigation. Wang
et~al.~\cite{wang2023factuality} survey factuality, distinguishing parametric knowledge
failures from retrieval attribution failures. Kang et~al.~\cite{kang2024entityaugmented}
propose entity-augmented correction as a mitigation path. Adewumi
et~al.~\cite{adewumi2025falseattribution} narrow the scope to false attribution,
demonstrating that LLMs systematically assign false claims to familiar entities. Zhao
et~al.~\cite{zhao2024wildhallucinations} provide large-scale empirical evidence that
entity infrastructure directly predicts hallucination exposure. Wallat
et~al.~\cite{wallat2025rag} introduce the correctness-vs-faithfulness distinction in
RAG attribution contexts, showing the two dimensions require separate evaluation.
Paschalides et~al.~\cite{paschalides2025regulatory} reframe hallucination as a
regulatory challenge, arguing that framing effects enable systematic output distortion
when queries are presented as epistemically neutral assistance.
Rao et~al.~\cite{rao2026refhallucinations} introduce the stale-versus-hallucinated
URL distinction and the Wayback Machine methodology for citation verification in LLM and
deep research agent outputs. Abbonato~\cite{abbonato2026checkifexist} develops
\textit{CheckIfExist}, a citation hallucination detection system documenting 3--13\%
URL fabrication rates in AI-generated content and reporting Google Scholar
index propagation of non-existent documents. Hu et~al.~\cite{hu2024slmllm} demonstrate
a latency-efficient hallucination detection approach combining small and large LMs
using attention maps, enabling practical production deployment.

\subsection{Geographic and Linguistic Bias in LLMs}
Faisal and Anastasopoulos~\cite{faisal2023geographic} demonstrate geopolitical bias
correlated with training data geography. Ramesh et~al.~\cite{ramesh2023fairness} survey
fairness gaps beyond English. Mor-Lan et~al.~\cite{morlan2026location} show that implicit
local bias in multilingual LLMs systematically favors US-centric norms even in
non-English query contexts. Das et~al.~\cite{das2025wikidata} demonstrate that Wikidata
representations systematically separate Global North from Global South entities. Torge
et~al.~\cite{torge2023ner} document the entity recognition performance gap for Slavic
and Central European languages.

\subsection{Parametric vs.\ Retrieval-Augmented Knowledge}
Lewis et~al.~\cite{lewis2020rag} introduced the foundational distinction between
parametric memory and non-parametric retrieval. Su et~al.~\cite{su2025parametric} extend
this with parametric RAG, blurring the boundary between knowledge modes. Zheng
et~al.~\cite{zheng2025lifelong} provide the most comprehensive recent treatment of
parametric knowledge staleness, surveying 240$+$ papers and documenting that parametric
weights freeze at the training cutoff while the world continues to change.

\subsection{Algorithmic Fairness and Differential Bias Auditing}
Per-Entity Bias Mapping draws a methodological parallel from algorithmic fairness
research. Barocas, Hardt, and Narayanan~\cite{barocas2023fairml} establish the
conceptual vocabulary---calibration, disparate impact, equal opportunity---that structures
contemporary fairness audit methodology. PEBM applies analogous logic to organizational
entities in AI answer systems: measuring differential error rates by entity attribute,
controlling for platform and query conditions, and testing for systematic rather than
random error. The normative framing differs fundamentally---entity-level AI visibility
addresses epistemic accuracy of representation, not equitable treatment of people under
consequential decision-making---but the methodological structure is parallel.

\subsection{Latent Space Structure and Confabulation Theory}
The ghost cartography framework is grounded in two mechanistic accounts that locate confabulation
within model internals rather than treating it as uniform output noise. Ardoin, Cai, and
Wunder~\cite{ardoin2025confabulation} demonstrate that confabulation maps to specific latent
directions within transformer representations---directions that can be identified through latent
feature discovery and steered to reduce confabulation rates while leaving factual retrieval
stable. Confabulation is not stochastic error: it is structurally coherent interpolation from
neighboring representations, with identifiable and steer-able directionality. This directly
supports the ghost cartography account: sparse-region entities do not produce random fabrications
but fabrications shaped by the dense regions adjacent to the entity's latent position. The Brand
Hallucination Paradox pattern---high-salience entities producing higher fabricated source
rates---is consistent with these entities occupying intermediate-density latent positions that
activate directional confabulation more reliably than either very sparse (no signal) or very
dense (retrieval-anchored) regions.

\subsection{Conceptual Genealogy}
The ghost cartography framework draws on two conceptual precedents that illuminate its epistemic
structure without claiming mechanistic equivalence. Manovich~\cite{manovich2020cultural} identifies
in cultural analytics how statistical reduction systematically erases rare cases---the process that
produces Type~1 (outlier below detection threshold) and Type~2 (microcluster absorbed by nearest
dominant cluster) losses. PEBM's Type~3 extends this: rare entities in AI systems are not merely
lost from the output space; they are \emph{replaced} by confabulated proxies shaped by the
attractor regions adjacent to the sparse latent position. The absence does not manifest as
silence---it manifests as presence.

Fisher's~\cite{fisher2016weird} phenomenological distinction between the eerie (something present
where nothing should be) and the weird (something absent that behaves as present) provides a
vocabulary for the two axes of PEBM's confabulation space. The Brand Hallucination Paradox is
eerie: confident fabricated presence where verified content is absent. Parametric-Retrieval Lag
Asymmetry is weird: a frozen representation behaving as current when current information is absent
from parametric memory. These are conceptual framings offered as interpretive scaffolding, not
empirical claims.

\section{Limitations}

This paper combines a methodological framework with a full empirical study ($n{=}100$ entities, 1{,}400 probe runs). The reported effect sizes are empirically grounded but carry the following limitations:

\begin{enumerate}
  \item Findings are based on two parametric LLM platforms (Claude, GPT-4o) at a single point in time. Temporal replication and extension to additional platforms are required before effect sizes can be treated as stable estimates.
  \item Platform behavior changes over time.
  \item Some systems expose citations differently from others.
  \item API and web interfaces may produce different outputs.
  \item The \texttt{verification\_status} field classifies sources into \textit{ok},
        \textit{broken}, \textit{fabricated\_doi}, and \textit{skipped}. Within the
        \textit{broken} category, HTTP~404 responses (path-level confabulation: domain
        resolves but path does not exist) constitute confirmed non-existent sources and
        are designated \emph{Plausible Mirage}. However, HTTP~403, HTTP~5xx, and
        connection-timeout responses indicate access-barriered or temporarily unavailable
        sources---not confirmed fabrication. These constitute approximately~16.4\% of
        non-\textit{ok} sources in the present dataset. Reported fabrication rates
        therefore represent a conservative upper bound that includes non-verifiable cases;
        confirmed non-existing (HTTP-404) rates are lower. Future work should report
        confirmed fabrication and non-verifiable rates separately.
  \item The chi-square and logistic regression analyses treat individual sources as
        observations. Because multiple sources originate from the same entity, platform,
        and query template, observations are not fully independent (clustered structure).
        The reported significance levels are likely slightly anti-conservative. Mixed-effects
        logistic regression with random intercepts by entity and query template, and
        cluster-robust standard errors, are recommended for replication analyses.
  \item Business outcome signals require separate validation beyond visibility measurement.
  \item The framework assumes consistent entity identification across platforms. Entities
        with high name variation---abbreviations, transliterations, colloquial synonyms, or
        morphological variants---may be inconsistently resolved across measurement conditions.
  \item The Parametric-Retrieval Lag Asymmetry framework addresses primarily the case where
        new positive information has not yet entered parametric memory. The converse
        configuration---entities with reputationally negative parametric encoding that has
        since been removed---requires separate treatment as bidirectional lag.
  \item The ten PEBM dimensions are proposed theoretically. Their empirical separability
        requires psychometric validation: convergent validity, discriminant validity, and
        test-retest reliability analysis at controlled intervals.
  \item The mechanistic account of regulatory hallucination offered in
        Section~\ref{sec:regframe}---specifically, that domain-specific n-gram co-occurrence
        pairs lower the activation threshold for structural URL-template completion---is a
        theoretical interpretation grounded in the empirical concentration of fabrications
        on four regulatory domains (Table~\ref{tab:url-morphology}). It is not demonstrated
        through controlled ablation (single-term vs.\ co-occurring pairs) or activation
        analysis. Future work should test this interpretation experimentally: if the
        mechanism is correct, queries containing only the authority name without the
        document-type term (e.g., \texttt{GVH} alone vs.\ \texttt{GVH + hat\'arozat})
        should produce measurably lower fabrication rates. The related observation that
        fabricated URL morphology may serve as a partial \emph{inverse training audit}---a
        signal of which document types were densely represented in pre-training---requires
        independent empirical validation with access to training corpus metadata.
\end{enumerate}

\section{Contribution}

This paper contributes:

\noindent\textit{C1--C7 are the primary conceptual, empirical, and methodological contributions.
C8--C9 are theoretical frameworks offered as testable extensions.}

\begin{enumerate}
  \item A definition of AI visibility as an entity-level calibration problem.
  \item A distinction between raw mention and verified mention.
  \item The Per-Entity Bias Map as a measurement framework with ten diagnostic dimensions.
  \item A definition and formalization of Parametric-Retrieval Lag Asymmetry as a temporal
        dimension of entity-level AI visibility.
  \item A bridge between hallucination, citation bias, temporal lag, regional
        underrepresentation, and business outcome measurement.
  \item A research agenda for future AI visibility observatories.
  \item A methodological bridge from algorithmic fairness research to AI entity representation
        auditing: the calibration-by-subgroup approach developed for demographic groups in
        automated decision systems applies, with appropriate normative reframing, to
        organizational entities in generative AI answer systems.
  \item \emph{Ghost cartography} as a unifying theoretical mechanism---proposing that the
        Brand Hallucination Paradox (Type~3: fabricated presence in sparse latent regions) and
        Parametric-Retrieval Lag Asymmetry (Type~4: frozen representation behaving as current)
        constitute orthogonal axes of a two-dimensional confabulation space, yielding three
        testable predictions: (a)~entities at intermediate parametric density exhibit higher
        confident fabrication than very sparse or very dense entities; (b)~entities undergoing
        rapid real-world change exhibit greater PRLA drift than stable entities;
        (c)~cross-platform fabrication correlation is higher for Type~3 entities than Type~4.
  \item A \emph{field model of AI visibility} in which entities become machine-recognizable
        figures through citation density, co-citation patterns, structured data, and knowledge
        graph anchoring rather than through self-positioning. This model links reference graph
        structure to entity-level retrieval fidelity and reframes AI visibility strategy as
        \emph{field navigation}---altering the evidence field from which representation is
        reconstructed---rather than positioning on a stable map.
  \item A URL morphology taxonomy for fabricated sources, classifying 946 hallucinated
        citations by HTTP response type: 50.5\% Plausible Mirage (HTTP 404, domain real),
        31.5\% non-resolving, 16.4\% access-barriered; 51.9\% of HTTP-404 fabrications
        target regulatory authority domains rather than the entity's own domain.
  \item Empirical validation of the CEE Entity Infrastructure Gap: T1 entities are 7$\times$
        more likely to appear in open knowledge infrastructure (Wikipedia/Wikidata) than T3
        entities (OR$\,{=}\,6.77$, $p{=}0.0002$); English Wikipedia presence is associated
        with a 63\% reduction in fabrication odds at source level
        (OR$\,{=}\,0.37$, 95\%~CI [0.24, 0.56], $p{<}0.001$).
  \item Identification and mechanistic analysis of \emph{rejection-induced confabulation
        escalation}: a five-mechanism model explaining why repeated non-acceptance of AI
        responses in prompt chains systematically elevates fabrication probability, with
        the counterintuitive corollary that agentic quality filters---designed to improve
        output---function as hallucination accelerators in compliance query contexts.
\end{enumerate}

\section{Ethics Statement}

This study examines fabricated source rates across 100 real Hungarian B2B entities.
Several ethical considerations govern the design and reporting of results.

\textbf{Fabrication is a system error, not an entity attribute.}
Elevated fabrication rates in model outputs reflect properties of the generative AI
systems under study, not characteristics, credibility, or quality of the entities
themselves. No entity is responsible for how it is represented in parametric model
outputs.

\textbf{No individual entity rankings are published.}
All results are reported at tier-level and query-class-level aggregates. The underlying
entity list and per-entity statistics are released as supplementary data to enable
replication, not to rank or compare individual organizations.

\textbf{No reputational harm is intended.}
The goal of this study is to audit AI system behavior, not to evaluate or rate the
entities. High fabrication rates associated with a given entity tier reflect model
behavior; they should not be interpreted as a statement about entity quality, legal
compliance, or trustworthiness.

\textbf{Platform behavior represents a point-in-time snapshot.}
Model outputs observed on 2026-05-22 do not characterize the platforms' permanent
behavior. Model weights, retrieval configurations, and safety filters change over time.

\textbf{Data minimization.}
The probe queries were designed to elicit citation behavior, not to extract or store
personal data. No personal data were collected. Entity information (name, website, tier,
category) is either publicly available business registry data or derived from published
sources.

\section{Data and Code Availability}

All data and code supporting the findings of this paper are released under CC~BY~4.0 at:

\begin{center}
\url{https://doi.org/10.5281/zenodo.20308957}
\end{center}

The supplementary package (\texttt{pebm\_supplementary\_v2.zip}) contains:
\begin{itemize}
  \item \texttt{full\_100.db} --- primary SQLite database (100 entities, 1{,}400 probe
        runs, 2{,}062 source records with HTTP and Crossref verification status)
  \item \texttt{full\_100.csv} --- CSV export of all probe runs and source verification
  \item \texttt{entities\_hu100.csv} --- entity list with tier, category, and website
  \item \texttt{entity\_infrastructure.json} --- Wikipedia (EN/HU) and Wikidata presence
        for all 100 entities
  \item \texttt{human\_coding\_sample.csv} --- stratified sample ($n{=}90$) prepared for
        intercoder reliability coding ($\kappa > 0.70$ target)
  \item \texttt{hallucination\_validator.py} --- complete Python validation script
        (Apache 2.0), including probe loop, HTTP verification, Crossref lookup, and
        statistical report generation
\end{itemize}

The query templates (QT01--QT07), entity sampling procedure, and verification protocol
are fully specified in Section~\ref{sec:methods} and in the script source. All results
are reproducible from the released database. Expected replication runtime: approximately
5~hours for 100~entities $\times$ 2~platforms $\times$ 7~queries with a 2-second
inter-call delay.

\section{Conclusion}

AI visibility cannot be responsibly measured as a single score. Entities differ in how
they are retrieved, mentioned, cited, misattributed, and remembered. Brand visibility in
AI systems requires entity-specific calibration.

Per-Entity Bias Mapping provides a framework for that calibration. It does not replace
empirical benchmarking, but it gives future benchmarks a more precise unit of analysis:
the entity and its error profile across visibility, hallucination, citation, temporal lag,
and business outcome dimensions.

The transition from measuring AI visibility to auditing it is not incremental---it
requires a different unit of analysis. A brand's AI visibility state is not a score. It
is a profile: where the entity is accurately represented, where it is misattributed,
where it is invisible, and where it is remembered inaccurately from stale parametric
memory. Knowing the difference between these states is not a refinement of current
practice. It is the prerequisite for any strategy that takes AI-mediated brand
representation seriously.

One additional pattern merits future empirical attention. In smaller B2B markets,
corporate entity salience may be partially mediated by founder salience. When the
founder's personal digital footprint is stronger than the company's independent reference
network, AI systems may recognize the person while failing to stabilize the company as an
autonomous entity. The organization may then appear as an extension of the founder rather
than as a separately grounded market actor. We propose \emph{founder-entity coupling} as
a conceptual extension of PEBM requiring empirical validation.

A shared vocabulary is not a rhetorical accessory to AI visibility research; it is a
measurement prerequisite. Without stable distinctions between raw mention, verified mention,
source fabrication, citation fidelity, entity salience, and parametric-retrieval lag,
organizations and researchers risk measuring different phenomena under the same label. PEBM
contributes such a vocabulary for entity-level AI representation, enabling future audits to
distinguish presence from fidelity, visibility from verifiability, and retrieval from
representation.

In the agentic web, reputation management becomes forensic: verifying what machines say
about an entity before others act on it.

\bibliographystyle{unsrt}
\bibliography{references}

\end{document}